\documentclass{ifacconf}

\usepackage{graphicx}      % include this line if your document contains figures
\usepackage{natbib}        % required for bibliography
\usepackage{mathtools}
\usepackage{amssymb}
\usepackage{amsmath}
\usepackage{algorithm}
\usepackage{algpseudocode}
\usepackage{algorithmicx}
\usepackage{url}
\usepackage{tikz}
\usepackage{pgfplots}
\usepgfplotslibrary{fillbetween}
\usepackage{natbib}
\usepackage{graphicx}
\usepackage{float}
\usepackage{array}
\usepackage{pgfgantt}
\usepackage{pdfpages}
\usepackage{lipsum}
\usepackage{algorithm}
\usepackage{upgreek}
\usepackage{booktabs}
\usepackage{filecontents}
\usepackage{pgfplotstable}
\pgfplotsset{compat=newest}
\usepackage{siunitx}

\DeclareMathAlphabet\mathbfcal{OMS}{cmsy}{b}{n}

\newcommand{\K}{\mathbf{K}}

\newcommand{\W}{\mathbf{W}}

\newcommand{\Wt}{\mathbfcal{W}}

\newcommand{\x}{\mathbf{x}}

\newcommand{\w}{\mathbf{w}}
\newcommand{\kk}{\mathbf{k}}
\newcommand{\y}{\mathbf{y}}

\newcommand{\id}{\mathbf{I}}
\newcommand{\Wd}{\W_{\setminus d}}
\newcommand{\bphi}{\boldsymbol{\Phi}}
\definecolor{blue}{RGB}{68,119,170}
\definecolor{cyan}{RGB}{102,204,238}
\definecolor{green}{RGB}{34,136,51}
\definecolor{yellow}{RGB}{238,204,102}
\definecolor{red}{RGB}{238,102,119}
\definecolor{purple}{RGB}{170,51,119}
\definecolor{orange}{RGB}{238,119,51}
\definecolor{grey}{RGB}{187,187,187}

\begin{document}
\begin{frontmatter}

\title{Projecting basis functions with tensor networks for Gaussian process regression} 

\author{Clara Menzen,} 
\author{Eva Memmel,} 
\author{Kim Batselier,} 
\author{Manon Kok}

\address{Delft Center for Systems and Control, TU Delft, Netherlands \{c.m.menzen,e.m.memmel,k.batselier,m.kok-1\}@tudelft.nl}

\begin{abstract}                
% Abstract of not more than 250 word
This paper presents a method for approximate Gaussian process (GP) regression with tensor networks (TNs).
A parametric approximation of a GP uses a linear combination of basis functions, where the accuracy of the approximation depends on the total number of basis functions $M$.
We develop an approach that allows us to use an exponential amount of basis functions without the corresponding exponential computational complexity.
The key idea to enable this is using low-rank TNs.
We first find a suitable low-dimensional subspace from the data, described by a low-rank TN. 
In this low-dimensional subspace, we then infer the weights of our model by solving a Bayesian inference problem. 
Finally, we project the resulting weights back to the original space to make GP predictions.
The benefit of our approach comes from the projection to a smaller subspace:
It modifies the shape of the basis functions in a way that it sees fit based on the given data, and it allows for efficient computations in the smaller subspace.
In an experiment with an 18-dimensional benchmark data set, we show the applicability of our method to an inverse dynamics problem.
%In our experiments, we first illustrate visually that projecting basis functions to a smaller subspace modifies the shape of the basis functions in a way that it sees fit based on the given data.
%Then, we show in a simulation that our method has superior performance to the competing methods when the data comes from a model where the weights are low-rank.

\end{abstract}


\end{frontmatter}
% the page limit is 6, but for review it can be up to 8
\section{Introduction}
A fundamental task in control is regression and a popular method of choice for regression with uncertainty bounds are Gaussian processes (GPs) \citep{Rasmussen2006}.
Being flexible function approximators, GPs are capable of using the information in data to learn rich representations and complex structures. 
In control, GPs have been applied for various problems, including model predictive control e.g.\ \cite{hewing2019cautious}, nonlinear state estimation e.g.\ \cite{berntorp2021online} and system identification e.g.\ \cite{chiuso2019system}.
For a tutorial for GPs in learning and control, we refer the reader to \cite{liu2018gaussian}.
Unfortunately, the appealing features of a GP come at a cost of poor scalability, with a computational complexity that grows cubically with the number of data points $N$, making GPs prohibitively expensive for large-scale data.

Among many efforts to make GPs feasible for large-scale data, see e.g.\ \cite{quinonero2005unifying,liu2020gaussian}, the work by \cite{HilbertGP} approximates the kernel function as a truncated series of inner products of basis functions. 
In this way, the GP model becomes parametric where each function value is computed as a linear combination of basis functions.
Predictions for unseen inputs can be made with a cost of $\mathcal{O}(NM^2)$, where $M$ is the number of basis functions that defines the quality of the kernel function approximation.
Consequently, in cases when $M$ needs to be chosen largely to achieve a good approximation, the complexity that is dominated by $M$ can become large.
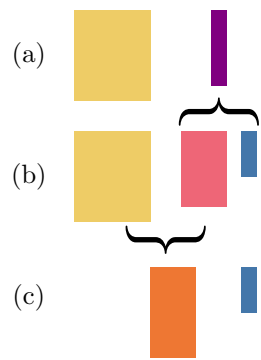
\begin{figure}
    \centering
    \begin{tikzpicture}
\node [rotate=90,scale=3] at (-1.7,8.06) {\}};
\node [rotate=-90,scale=3] at (-2.4,6.36) {\}};
\node at (-4.2,8.8) {(a)};
\node at (-4.2,7.2) {(b)};
\node at (-4.2,5.6) {(c)};
\fill [yellow]  (-3.6,9.4) rectangle (-2.6,8.2);
\fill [violet]  (-1.8,9.4) rectangle (-1.6,8.4);
\fill [yellow]  (-3.6,7.8) rectangle (-2.6,6.6);
\fill [red]  (-2.2,7.8) rectangle (-1.6,6.8);
\fill [blue]  (-1.4,7.8) rectangle (-1.2,7.2);
\fill [orange] (-2.6,6) rectangle (-2,4.8);
\fill [blue] (-1.4,6) rectangle (-1.2,5.4);
\end{tikzpicture}
    \caption{Illustration of the fundamental idea of our proposed method. (a) GP approximation in terms of a parametric model with basis functions (yellow) and corresponding weights (violet). (b) Projection matrix (red) that projects the basis functions to a smaller subspace where a smaller weight vector (blue) lives. (c) `New' parametric model with projected basis functions (orange) and a smaller weight vector (blue).}
    \label{fig:twostepalgo}
\end{figure}
In this work, we overcome this limitation with promising tools from multilinear algebra: tensor networks (TNs).
TNs, as e.g.\ described in \cite{Kolda2009}, are networks of multidimensional arrays, also called tensors, that can approximate data or models in a compressed format while preserving the essence of the information. 
To the best of the authors' knowledge, there is little literature that applies TN in the framework of GPs.
Existing work uses TNs for scalability and/or model interpretability and expressiveness - see e.g.\ \cite{izmailov2018scalable, pmlr-v179-kirstein22a, konstantinidis2021kernel}.
We apply TNs in the framework of \cite{HilbertGP} to perform GP regression with a computational complexity of $\mathcal{O}(R^4M_d^2N)$, where $R$ is called the rank of a TN, and where $M_d$ is the number of basis functions for one dimension, such that $M_d$ is much smaller than the total number of basis functions $M$.

Our contribution is inspired by \cite{calandra2016manifold}.
In their work, a manifold is learned to transform the inputs into a low-dimensional space where GP regression is performed.
In this way, functions that e.g.\ violate the smoothness assumption can still be described well by a GP in the low-dimensional space.
We take inspiration from this idea in terms of how we transform the parametric model that approximates the GP (Fig.~\ref{fig:twostepalgo}~(a)).
As a first step, the weights are split into a tall matrix (red) and a vector that has fewer components than the original weight vector (blue) (Fig.\ \ref{fig:twostepalgo}~(b)). 
This is achieved by computing the weight vector as a low-rank TN with the well-known alternating linear scheme (ALS).
Then, the tall projection matrix (red) is absorbed into the matrix containing the collection of basis functions (Fig.\ \ref{fig:twostepalgo}~(c)).
We interpret this step as projecting the basis functions into a smaller subspace.
The second step is to solve a Bayesian inference problem with the projected model to find a posterior distribution of the smaller weight vector.
Finally, this distribution can be projected back to the original space to do approximate GP predictions.

\section{Background}
In this section, we briefly review the necessary background on GP regression and the reduced-rank approximation introduced by \cite{HilbertGP}, as well as kernel machines, where the model weights are modeled as a TN.

\subsection{Gaussian process regression}
Given a data set of $N$ observations, $\x_1,\x_2,\dots,\x_N$ and $\y=y_1,y_2,\dots,y_N$, the GP model is given by
\begin{equation}
\begin{aligned}
    y_n &= f(\x_n) + \epsilon_n,\quad\epsilon_n\sim\mathcal{N}(0,\sigma_y^2)\\
    f(\x) &\sim \mathcal{GP}(m(\x),\kappa(\x,\x')),
\end{aligned}
    \label{eq:model}
\end{equation}
where a mean function $m(\cdot)$ and covariance function also called kernel function $\kappa(\cdot,\cdot)$ define the GP, $\sigma_y^2$ denotes the variance of the measurement noise $\boldsymbol\epsilon$, that is assumed to be zero-mean Gaussian and i.i.d., and $\x,\x^\prime\in\mathbb{R}^{1\times D}$ are $D$-dimensional inputs. 

In GP regression, the well-known prediction equations for an unseen input $\x_*\in\mathbb{R}^{1\times D}$ are given by
\begin{equation}\begin{aligned}
f_* \mid \y  &\sim \mathcal{N}\left(\mathbb{E}\left(f_*\right), \mathbb{V}\left(f_*\right)\right), \text { with } \\
\mathbb{E}\left(f_*\right) & =\kk_*^\top\left[\K+\sigma_y^2 \id_N \right]^{-1} \y\;\; \text{and} \\
\mathbb{V}\left(f_*\right) &=k_{**}-\kk_*^\top\left[\K+\sigma_y^2 \id_N\right]^{-1} \kk_*, \label{eq:posterior_mP}
\end{aligned}\end{equation}
where the entries of the kernel matrices are computed with
\begin{equation*}
    [\K]_{nn^\prime} = \kappa(\x_n,\x_{n^\prime}),\;[\kk_*]_n = \kappa(\x_n,\x_*)\;\text{and}\;k_{**} = \kappa(\x_*,\x_*).
\end{equation*}

The required matrix inversion in \eqref{eq:posterior_mP} is a GP's computational bottleneck of $\mathcal{O}(N^3)$ when $N$ is large.

\subsection{Hilbert space methods for reduced-rank GP regression}
\cite{HilbertGP} approximate the kernel function by a truncated series of inner products between basis functions.
More specifically, given input vectors that lay centered in a domain confined by a $D$-dimensional hyperbox $[-L_1,L_1]\times [-L_2,L_2]\times \dots \times [-L_D,L_D]$, the kernel function approximation is given by
\begin{equation}
    \kappa(\x,\x^\prime) \approx \sum_{m=1}^M S([\boldsymbol\lambda]_m) \boldsymbol\phi_m(\x)\boldsymbol\phi_m^\top(\x^\prime).
    \label{eq:truncseries}
\end{equation}
The function $\boldsymbol\phi_m(\x)$ is the $m$th eigenfunction of the negative Laplace operator on the confined hyperbox domain, subject to boundary conditions, such as Dirichlet.

The basis functions $\boldsymbol\phi_m(\x)$ are multivariate  basis functions, computed as a product $D$ univariate eigenfunctions, where the $(m_d)$th entry is computed with
\begin{equation}
    \left[\boldsymbol\phi\left([\x_n]_d\right)\right]_{m_d} = \frac{1}{\sqrt{L_d}}\sin\left(\pi m_d\frac{[\x_n]_d+L_d}{2L_d} \right).
    \label{eq:mdeigenfunc}
\end{equation}
The entry $[\x_n]_d$ is the $d$th dimension of the $n$th input and it is $\boldsymbol\phi\left([\x_n]_d\right)\in\mathbb{R}^{M_d}$, where $M_d$ is the number of basis functions for the $d$th dimension.
The corresponding eigenvalue is denoted by $[\boldsymbol\lambda]_m$.

The eigenvalue corresponding to the $m$th eigenfunction is computed as a sum of $D$ terms, where each term is computed with 
\begin{equation}
    [\boldsymbol\lambda]_{m_d} = \left(\frac{\pi {m_d}}{2L_d}\right)^2.
    \label{eq:mdeigenval}
\end{equation}
The function $S(\cdot)$ denotes the spectral density of the Gaussian kernel, computed dimension-wise with
\begin{equation}
\begin{aligned}
     S(\sqrt{[\boldsymbol\lambda]_{m_d}})= \sigma_f^2\sqrt{2\pi}\ell\exp\left(-\frac{\ell^2}{2}[\boldsymbol\lambda]_{m_d}\right),
     \label{eq:spectralden}
\end{aligned}
\end{equation}
where $\sigma_f^2$ and $\ell$ denote the hyperparameters, i.e., signal variance and length scale, respectively.

Given \eqref{eq:truncseries}, a reduced-rank approximation of the kernel matrix is given by
\begin{equation}
    \K \approx \bphi\boldsymbol\Lambda\bphi^\top = (\bphi\boldsymbol\Lambda^\frac{1}{2})(\boldsymbol\Lambda^\frac{1}{2}\bphi^\top),
    \label{eq:kernapprox}
\end{equation}
where $\bphi\in\mathbb{R}^{N\times M}$ contain the basis functions, $\boldsymbol\Lambda\in\mathbb{R}^{M\times M}$ contain the leading $M$ eigenvalues and $\boldsymbol\Lambda^\frac{1}{2}$ is the Cholesky factor of $\boldsymbol\Lambda$. 

With the given kernel approximation, the GP can be written as a parametric model given by
\begin{equation}
    \y = \bphi\w + \boldsymbol\epsilon,\;\; \boldsymbol\epsilon\sim\mathcal{N}(\boldsymbol0,\sigma_y^2\id_N),
    \label{eq:parammodel}
\end{equation}
where $\w\sim\mathcal{N}(\boldsymbol0,\boldsymbol\Lambda)$ are the model weights.
The prediction equations for this reduced-rank approach are given by
\begin{equation}
    \begin{aligned}
    \mathbb{E}(f_*) &= \boldsymbol\phi_*\underbrace{(\bphi^\top\bphi + \sigma_y^2\boldsymbol\Lambda^{-1})^{-1}\bphi^\top\y}_{\mathbb{E}(\w\mid\y)}\\
    \mathbb{V}(f_*) &= \boldsymbol\phi_*\underbrace{\sigma_y^2(\bphi^\top\bphi + \sigma_y^2\boldsymbol\Lambda^{-1})^{-1}}_{\mathbb{V}(\w\mid\y)}\boldsymbol\phi_*^\top,
    \end{aligned}
    \label{eq:post}
\end{equation}
where $\w\mid\y$ denotes the posterior distribution of the weights.

In terms of computational complexity, the reduced-rank approach requires $\mathcal{O}(M^3)$ for hyperparameter training and $\mathcal{O}(NM^2)$ for inference.
The number of basis functions $M$ needs to be chosen based on multiple criteria: the size of the domain, the required accuracy to approximate the kernel matrix, and the dimensionality of the problem.

\subsection{Kernel machines in TN format}
In the context of kernel machines, \cite{stoudenmire2016supervised,batselier2017tensor,wesel2021large}, model the weights of the parametric model as a low-rank TN, such that it is
\begin{equation}
\begin{aligned}
        \y &= \bphi\w+ \boldsymbol\epsilon\\
        \text{s.t.} \; \w \; &\text{being a low-rank TN},
        \label{eq:lowrankmodel}
\end{aligned}
\end{equation}
where the model weights are not treated as a random variable, but as deterministic. 
Furthermore, the matrix $\bphi$, containing the basis functions, is written as a row-wise Khatri-Rao product given by
\begin{equation}
    \begin{aligned}
    \boldsymbol\Phi &= 
    \begin{bmatrix}
    \boldsymbol\phi([\x_1]_1) \otimes \cdots\otimes \boldsymbol\phi([\x_1]_d) \otimes \cdots \otimes\boldsymbol\phi([\x_1]_D) \\
    \vdots\\
    \boldsymbol\phi([\x_n]_1) \otimes\cdots\otimes \boldsymbol\phi([\x_n]_d) \otimes \cdots \otimes\boldsymbol\phi([\x_n]_D)\\
    \vdots\\
    \boldsymbol\phi([\x_N]_1) \otimes\cdots\otimes \boldsymbol\phi([\x_N]_d) \otimes \cdots \otimes\boldsymbol\phi([\x_N]_D)
    \end{bmatrix},
    \end{aligned}
    \label{eq:prodkern}
\end{equation}
such that the $n$th row is a vector computed as a Kronecker product of $D$ vectors $\boldsymbol\phi([\x_n]_1),\dots,\boldsymbol\phi([\x_n]_d),\dots,\boldsymbol\phi([\x_n]_D)$. 
The $d$th vector $\boldsymbol\phi([\x_n]_d)\in\mathbb{R}^{1\times M_d}$ contains $M_d$ basis functions where the subscript $d$ denotes the $d$th dimension. 

Assuming the same number of basis functions in each dimension, i.e., $M_1=M_2=\dots=M_D$, the storage cost of $\bphi$ is in principle $\mathcal{O}(NM_d^D)$. 
However, in practice there is no need to compute $\bphi$ explicitly, because it can be stored as $D$ matrices of size $N\times M_d$. 
In this way, the storage cost of $\bphi$ is linear in $D$, i.e., $\mathcal{O}(NDM_d)$.

\cite{wesel2021large} approximate the Gaussian kernel with the structure in \eqref{eq:prodkern}, by computing the basis functions as \eqref{eq:mdeigenfunc} weighted with \eqref{eq:mdeigenval}.

To solve \eqref{eq:lowrankmodel}, \cite{stoudenmire2016supervised,batselier2017tensor} model $\w$ as a tensor train (TT) decomposition \cite{Oseledets2011} and \cite{wesel2021large} as a CANDECOMP/PARAFAC decomposition \citep{Kolda2009}.
Since this paper focuses on the TT decomposition, we will consider this case in the following.
A weight vector modeled as a TT is given in terms of three-dimensional tensors, called TT-cores denoted by $\Wt^{(1)},\Wt^{(2)},\dots,\Wt^{(D)}$, where $\Wt^{(d)}\in\mathbb{R}^{R_dM_dR_{d+1}}$ and $R_1,R_2,\dots,R_{D+1}$ are the ranks of the decomposition. 
The ranks determine the accuracy of the representation, as well as the complexity of computations with a TT.
According to the definition of the TT decomposition, the tensor $\Wt$ representing $\w$ can be computed element-wise with
\begin{equation*}
[\Wt]_{i_1i_2\dots i_D}=\sum_{r_1=1}^{R_1} \cdots \sum_{r_{D+1}=1}^{R_{D+1}} \left[\Wt_1\right]_{r_1i_1r_2}\cdots\left[\Wt_D\right]_{r_Di_Dr_{D+1}},
\end{equation*}
where $R_1=R_{D+1}=1$.

Because of the multilinear nature of the TT decomposition, a vector represented as a TT can be written as a function that is linear with respect to a specific TT-core.
This is achieved by computing a matrix $\Wd\in\mathbb{R}^{M\times R_dM_dR_{d+1}}$ from all TT-cores except the $d$th, such that $\w$ is modeled as 
\begin{equation}
    \w = \Wd\w^{(d)},
    \label{eq:wTT}
\end{equation}
where $\w^{(d)}\in\mathbb{R}^{R_dM_dR_{d+1}}$ is the vectorized $d$th TT-core. 

Exploiting the multilinearity of the TT decomposition, the regularized least square problem to find the model weights is given by
\begin{equation}
    \min_{\w^{(d)}}\| \y-\bphi\boldsymbol\Wd\w^{(d)} \|_2^2 + \lambda \|\Wd\w^{(d)}\|_2^2,
    \label{eq:min}
\end{equation}
where $\lambda$ denotes the regularization parameter.

Equation \eqref{eq:min} is solved by applying the well-known alternating linear scheme (ALS).
The ALS is an iterative block coordinate descent method that updates one TT-core at a time while assuming the other TT-cores to be known and fixed. 
One so-called sweep of the ALS is updating all TT-cores once by solving \eqref{eq:min} sequentially. 
After multiple sweeps, when a convergence criterion is met, a TT representation of the model weights $\w$ is found.

In the ALS, there is a simple way to avoid that the conditioning of the subproblem becomes worse than the overall problem: keeping the TT in site-$d$-mixed canonical format \cite[p.113]{holtz2012alternating}. 
In this format, the TT-cores are computed in a way such that it is
\begin{equation}
\Wd^\top\Wd=\id_{R_dM_dR_{d+1}}.     
\end{equation}

\section{Methods}
\label{sec:methods}
In our method, we compute the mean and covariance of a weight vector with a computational complexity of $\mathcal{O}(R^4M_d^2N)$, where $R$ can be assumed to be small and $M_d\ll M$.
We combine the model of the kernel machine \eqref{eq:lowrankmodel} that models the weight as a low-rank TN, and the probabilistic parametric model \eqref{eq:parammodel}, that approximated a GP.
In this way, we can exploit the TN structure that allows for efficient computation and at the same time take prior knowledge about the weights into account.
Subsequently, we compute a posterior distribution for the weights in a TN framework, allowing for both a mean and covariance estimate for the predictions.

In this context, we consider an equivalent way of writing model \eqref{eq:parammodel} and \eqref{eq:post}, given by
\begin{equation}
    \y = \bphi\boldsymbol\Lambda^\frac{1}{2}\w + \boldsymbol\epsilon, \;\; \boldsymbol\epsilon\sim\mathcal{N}(\boldsymbol0,\sigma_y^2\id_N),\;\; \w\sim\mathcal{N}(\boldsymbol0,\id)
    \label{eq:weightedmodel}
\end{equation}
and
\begin{equation}
    \begin{aligned}
    \mathbb{E}(f_*) &= \boldsymbol\phi_*\boldsymbol\Lambda^\frac{1}{2}\underbrace{(\bphi^\top\boldsymbol\Lambda\bphi + \sigma_y^2\id_M)^{-1}\boldsymbol\Lambda^\frac{1}{2}\bphi^\top\y}_{\mathbb{E}(\w\mid\y)}\\
    \mathbb{V}(f_*) &= \boldsymbol\phi_*\boldsymbol\Lambda^\frac{1}{2}\underbrace{\sigma_y^2(\bphi^\top\boldsymbol\Lambda\bphi + \sigma_y^2\id_M)^{-1}}_{\mathbb{V}(\w\mid\y)}\boldsymbol\Lambda^\frac{1}{2}\boldsymbol\phi_*^\top,
    \end{aligned}
\end{equation}

Combining \eqref{eq:weightedmodel} with \eqref{eq:lowrankmodel}, in our method we consider the model 
\begin{equation}
    \y = \bphi \boldsymbol\Lambda^\frac{1}{2} \Wd\w^{(d)} + \boldsymbol\epsilon,\;\; \boldsymbol\epsilon\sim\mathcal{N}(\boldsymbol0,\sigma_y^2\id_N),
    \label{eq:paramm}
\end{equation}
where $\Wd$ is deterministic and $\w^{(d)}$ has a prior given by
\begin{equation}
\w^{(d)}\sim \mathcal{N}(\boldsymbol0,\id_{R_dM_dR_{d+1}}).
% \Wd^\top(\Wd\w^{(d)})\sim \mathcal{N}(\boldsymbol0,\Wd^\top\Wd) \text{projection down}
\label{eq:priorwd}
\end{equation}
The matrix $\bphi$ is given in the format \eqref{eq:prodkern} and the diagonal matrix is given as a Kronecker product over dimensions 
\begin{equation}
    \begin{aligned}
    \boldsymbol\Lambda = \boldsymbol\Lambda^{(1)} \otimes \boldsymbol\Lambda^{(2)} \otimes \cdots \otimes \boldsymbol\Lambda^{(D)},
    \end{aligned}
    \label{eq:lambdakron}
\end{equation}
where each entry of $\boldsymbol\Lambda^{(d)}$ is computed with \eqref{eq:spectralden}.

In order to approximate a posterior distribution for the weights $\w\mid\y$ and do predictions for unseen input locations, our proposed algorithm, summarized in Alg.\ \ref{alg:algi}, consists in two consecutive steps. 
First, we compute $\Wd$, a projection matrix that projects the weight vector into a smaller subspace, where we compute the posterior distribution of $\w^{(d)}\mid\y$.
Second, $\w^{(d)}\mid\y$ is projected to a posterior distribution $\w\mid\y$, to enable predictions for unseen inputs.

More specifically, the projection matrix $\Wd$ is computed with the regularized ALS (line 1 of Alg.\ \ref{alg:algi}).
After a TT representation of the weight vector $\w$ with TT-cores $\Wt^{(1)},\Wt^{(2)},...,\Wt^{(D)}$ is found, $\Wd$ is computed from all the TT-cores except the $d$th, such that it is $\w=\Wd\w^{(d)}$. 
%with
%\begin{equation}
%    \Wd = \W_{i>d}\otimes\id_{M_d}\otimes\W_{i<d}^\top,
%    \label{eq:Wd}
%\end{equation}
%where $\W_{i>d}\in\mathbb{R}^{M_{d+1}\dots M_D\times R_{d+1}}$ and %$\W_{i<d}\in\mathbb{R}^{M_{1}\dots M_{d-1}\times R_{d}}$ are matrices computed from %the TT-cores of the right and left side of the $d$th TT-core, such that it is %$\w=\Wd\w^{(d)}$. 

Exploiting the TT structure of $\w$ as well as the Khatri-Rao structure of $\bphi$ and the Kronecker structure in $\boldsymbol\Lambda$, the matrix-matrix-multiplications $\bphi\boldsymbol\Lambda^\frac{1}{2}\Wd$ in \eqref{eq:min} are performed without explicitly constructing the matrices.
Instead, the multiplication is performed in an efficient way, for each dimension separately.

After the computation of the projection matrix $\Wd$, a Bayesian update to compute $\w^{(d)}\mid\y$ is performed (line~2 of Alg.\ \ref{alg:algi}).
Considering the prior distribution given in \eqref{eq:priorwd}
the mean and covariance of the posterior distribution $\w^{(d)}\mid\y$ is given by
\begin{equation}
    \begin{aligned}
        \mathbb{V}(\w^{(d)}\mid\y) &= \sigma_y^2\left[\Wd^\top\boldsymbol\Phi^\top\boldsymbol\Lambda \boldsymbol\Phi\Wd + \sigma_y^2 \id_{R_dM_dR_{d+1}} \right]^{-1} \\
        \mathbb{E}(\w^{(d)}\mid\y) &= \mathbb{V}(\w^{(d)}\mid\y)\sigma_y^{-2}\Wd^\top\boldsymbol\Lambda^\frac{1}{2}\boldsymbol\Phi^\top\y.  
        \label{eq:wd}
    \end{aligned}
\end{equation}

The computation of $\Wd^\top\boldsymbol\Phi^\top\boldsymbol\Lambda \boldsymbol\Phi\Wd$ in \eqref{eq:wd} has a computational complexity of $\mathcal{O}(R^4M_d^2N)$, where $R$ denotes the maximum rank of the TT and assuming $M_1=M_2=\dots=M_D$.
The complexity depends on $R^4$, as a consequence the ranks of the TT need to be small to achieve a significant speed-up. 
Since the choice of the ranks also influences the accuracy of the approximation, a trade-off needs to be made. 

After computing the posterior distribution $\w^{(d)}\mid\y$, the projection matrix is used to project it to $\w\mid\y$ (line 3 of Alg.\ \ref{alg:algi}) with
\begin{equation}
\begin{aligned}
\mathbb{E}(\w\mid\y) &= \Wd\mathbb{E}(\w^{(d)}\mid\y)\\
\mathbb{V}(\w\mid\y) &= \Wd\mathbb{V}(\w^{(d)}\mid\y)\Wd^\top.
\label{eq:proj}
\end{aligned}
\end{equation}
Since $\Wd$ is a tall matrix and $\mathbb{V}(\w^{(d)}\mid\y)$ is square, it can be seen that the posterior covariance $\mathbb{V}(\w\mid\y)$ is approximated as a low-rank matrix, where the rank corresponds to the number of elements in $\w^{(d)}$.

Finally, the obtained mean and covariance for $\w\mid\y$ are used to make predictions for unseen input locations (line 4 of Alg.\ \ref{alg:algi}) with
\begin{equation}
    \begin{aligned}
    \mathbb{E}(f_*) &=  \boldsymbol\phi_*\boldsymbol\Lambda^\frac{1}{2}\mathbb{E}(\w\mid\y)\\
    \mathbb{V}(f_*) &=  \boldsymbol\phi_*\boldsymbol\Lambda^\frac{1}{2}\mathbb{V}(\w\mid \y)\boldsymbol\Lambda^\frac{1}{2}\boldsymbol\phi_*^\top.
    \end{aligned}
    \label{eq:pred_hilbertgp_w}
\end{equation}
In practice, $\boldsymbol\phi_*\boldsymbol\Lambda^\frac{1}{2}$ is not computed explicitly. 
Instead, the product $\boldsymbol\phi_*\boldsymbol\Lambda^\frac{1}{2}\Wd$ is computed efficiently exploiting the structure of the matrices, resulting in a vector of size $1\times R_dM_dR_{d+1}$.

\begin{algorithm}
\caption{Gaussian process regression with projected basis functions}\label{alg:algi}
\begin{algorithmic}[1]
\Require Collections of basis functions $\bphi$ and $\boldsymbol\phi_*$, matrices $\boldsymbol\Lambda^{(1)}$,...,$\boldsymbol\Lambda^{(D)}$, measurements $\y$, noise variance $\sigma_y^2$
\Ensure Predictive distribution $f_*\mid\y$
\State Compute $\Wt^{(1)},\Wt^{(2)},...,\Wt^{(D)}$ with ALS, then $\Wd$ from $\Wt^{(1)},...,\Wt^{(d-1)},\Wt^{(d+1)},...,\Wt^{(D)}$.
\State Compute $\w^{(d)}\mid\y$ with Bayesian inference in projected subspace with equation \eqref{eq:wd}.
\State Project $\w^{(d)}\mid\y$ to $\w\mid\y$ with \eqref{eq:proj}.
\State Predict in unseen location using \eqref{eq:pred_hilbertgp_w}.
\end{algorithmic}
\end{algorithm}

\section{Numerical Experiments}
\begin{figure*}
\begin{minipage}{0.45\textwidth}
\centering
    \includegraphics[width=\textwidth]{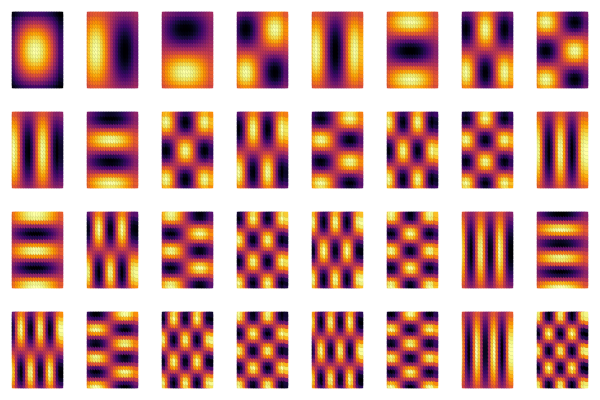}
    \caption{Leading 32 basis functions computed with \cite{HilbertGP}, sorted by the descending magnitude of the eigenvalue spectrum.}
    \label{fig:bf1}
\end{minipage}\hfill
\begin{minipage}{0.45\textwidth}
\centering
    \includegraphics[width=\textwidth]{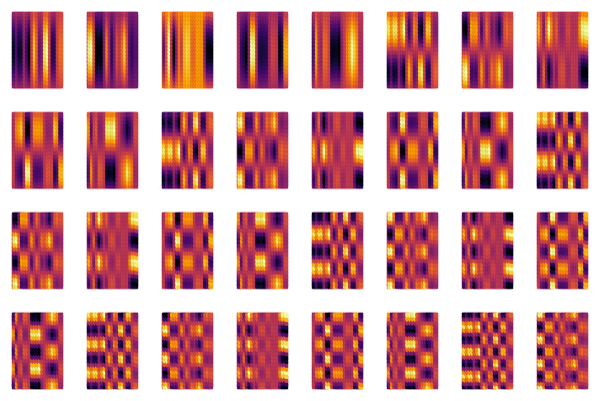}
    \caption{Subset of $R^2M_2=1000$ projected basis functions stored in columns of $\bphi\Wd\in\mathbb{R}^{N\times R^2M_2}$ computed with Alg.~\ref{alg:algi}. }
    \label{fig:bf2}
\end{minipage}
\end{figure*}
\label{sec:numexp}
All simulations (section \ref{sec:bfana} and \ref{sec:predsynth}) are performed on a Lenovo computer with processor Intel(R) Core(TM) i7-8650U CPU @ 2.11 GHz and 16GB of RAM. 
The experiment in section \ref{sec:arm} is performed on an AMD EPYC 7252 8-Core Processor 3.10 GHz with 256 GB of RAM.
The implementation of the method, simulations, and experiments can be found at 
\url{https://github.com/clarazen/ProjectedBasisFunctions} commit \texttt{a03b47b}.

For all simulations, synthetic data is generated with the model \eqref{eq:paramm}.
The matrix $\bphi$ is computed as \eqref{eq:prodkern} with \eqref{eq:mdeigenfunc} and an input that is specified for each experiment in the corresponding subsection.
The matrix $\Wd$, which is usually computed in the first step of Alg.\ \ref{alg:algi}, for simplicity, is initialized randomly, then its columns are orthogonalized.
The size of $\Wd$, i.e. $M\times~RM_dR$, depends on both $d$, specified in the subsections of the experiments, as well as on the TT-ranks which are assumed to be all the same $R=R_1=\dots=R_D$.

\subsection{Basis function analysis}
\label{sec:bfana}
In our first simulation, we investigate, how the dominant basis functions computed in \cite{HilbertGP} compare visually to the same basis functions that are projected to a smaller subspace with Alg.\ \ref{alg:algi}.
In this context, we generate $N=400$ input locations on a regular 2D grid confined by $[-1,1]\times[-1,1]$ using the hyperparameters $\ell^2=0.01$ and $\sigma_f^2=1.0$ and $R=5$.
Then, we compute $\bphi\boldsymbol\Lambda^\frac{1}{2}\in\mathbb{R}^{N\times M_1M_2}$ with $M_d=40$.
Figure \ref{fig:bf1} shows the leading 32 of the 1600 basis functions in descending order.
As can be seen in the subfigures, the dominating basis functions are low-frequent with an increasing frequency the lower the eigenvalue to the corresponding basis function.
Figure \ref{fig:bf2} shows the projected basis functions or columns of the matrix $\bphi\boldsymbol\Lambda^\frac{1}{2}\W_{\setminus 2}$.
Comparing both figures gives an insight into how Algorithm \ref{alg:algi} works: 
The projection changes the shape of the basis functions in a way that it sees fit based on $\W_{\setminus 2}$ which is generally computed from the given data. 

\subsection{Prediction accuracy on synthetic data}
\label{sec:predsynth}
In our second simulation, we investigate how our method performs in terms of prediction accuracy on data generated from \eqref{eq:paramm}, and compare it to the full GP, as well as the reduced-rank approach by \cite{HilbertGP} (Hilbert-GP).
Given an input with $D=3$ that is randomly sampled from a box confined by $[-1,1]\times[-1,1]\times[-1,1]$, we sample $N=4000$ training observations and $N_*=1000$ validation observations with hyperparameters $\ell^2=0.02$ and $\sigma_f^2=1.0$.
We compute $\W_{\setminus 2}\in\mathbb{R}^{RM_2R}$ with $R=1,5,10,20$ as described in the introduction of section \ref{sec:numexp}, and choose $\sigma_y^2$ such that the signal-to-noise ratio is $10\;\mathrm{dB}$.
For the ALS, we choose the same ranks with which we create the data, as we assume them to be known in advance.
For the comparison with Hilbert-GP, we choose the dominating $R^2M_2$ basis functions.

\begin{figure}
    \centering
    % Recommended preamble:
\begin{tikzpicture}
\begin{axis}[xmajorgrids, ymajorgrids, xlabel={TT-ranks}, ylabel={RMSE}, xtick={1,2,3,4}, xticklabels={1,5,10,20}, style = {very thick}, legend style={at={(0.5,0.4)},anchor=west}, ymode=log,height=5.1cm, width=\columnwidth]
    \addplot[color={red}, mark={x}, error bars/y dir=both, error bars/y explicit]
        coordinates {
            (1.0,0.01410697453160865) +- (0,0.005287304008331177)
            (2.0,0.04916051120989571) +- (0,0.003928753531868679)
            (3.0,0.08825318266587927) +- (0,0.00507586667052358)
            (4.0,0.15953848262399364) +- (0,0.008124867873886403)
        }
        ;
    \addlegendentry {Full GP}
    \addplot[color={blue}, mark={x}, error bars/y dir=both, error bars/y explicit]
        coordinates {
            (1.0,0.03475561678229831) +- (0,0.019167984190144437)
            (2.0,0.0705370553262669) +- (0,0.00767804860109263)
            (3.0,0.08764309435206637) +- (0,0.004844152460928361)
            (4.0,0.1595381470880651) +- (0,0.00812485708202234)
        }
        ;
    \addlegendentry {Hilbert-GP}
    \addplot[color={green}, mark={x}, error bars/y dir=both, error bars/y explicit]
        coordinates {
            (1.0,0.0011991447136529237) +- (0,0.0005740787831675132)
            (2.0,0.024209141449643153) +- (0,0.002830868099943029)
            (3.0,0.08568418626432074) +- (0,0.0051559268788238)
            (4.0,0.15953814708806538) +- (0,0.008124857082022275)
        }
        ;
    \addlegendentry {Algorithm 1}
\end{axis}
\end{tikzpicture}
    \caption{RMSE on validation data for data generated with model \eqref{eq:paramm} with varying ranks. Mean and standard deviation plotted from 10 different runs.}
    \label{fig:reerr}
\end{figure}
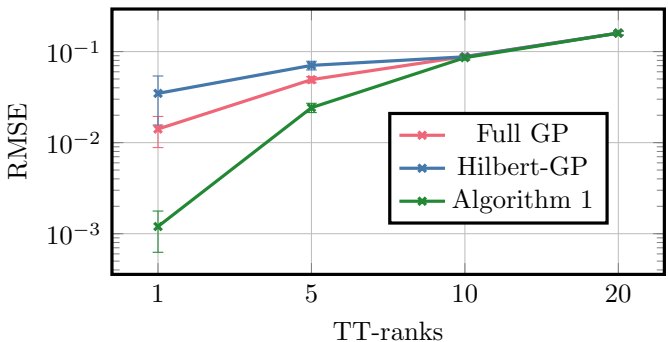

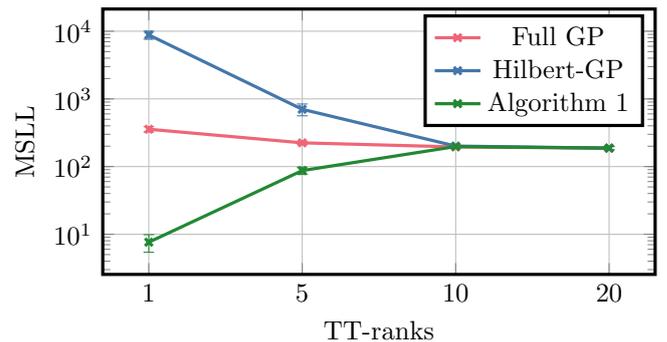
\begin{figure}
    \centering
    % Recommended preamble:
\begin{tikzpicture}
\begin{axis}[xmajorgrids, ymajorgrids, xlabel={TT-ranks}, ylabel={MSLL}, xtick={1,2,3,4}, xticklabels={1,5,10,20}, style = {very thick},ymode=log, ymode=log,height=5.1cm, width=\columnwidth]
    \addplot[color={red}, mark={x}, error bars/y dir=both, error bars/y explicit]
        coordinates {
            (1.0,356.2376937320522) +- (0,33.285657321902086)
            (2.0,224.06939421652396) +- (0,12.266663215432652)
            (3.0,195.0784016783536) +- (0,8.69688552219202)
            (4.0,187.27212138937026) +- (0,10.305102175684084)
        }
        ;
    \addlegendentry {Full GP}
    \addplot[color={blue}, mark={x}, error bars/y dir=both, error bars/y explicit]
        coordinates {
            (1.0,8821.689934699241) +- (0,1209.1130649751483)
            (2.0,701.9228197130926) +- (0,139.25169447714913)
            (3.0,200.87328432410843) +- (0,7.97372283109239)
            (4.0,187.28025410457832) +- (0,10.30571140066462)
        }
        ;
    \addlegendentry {Hilbert-GP}
    \addplot[color={green}, mark={x}, error bars/y dir=both, error bars/y explicit]
        coordinates {
            (1.0,7.62748598485152) +- (0,2.2210276279764782)
            (2.0,87.23236630821614) +- (0,10.834531218622825)
            (3.0,197.50364743919326) +- (0,6.6419145501273)
            (4.0,187.28025410457886) +- (0,10.305711400664828)
        }
        ;
    \addlegendentry {Algorithm 1}
\end{axis}
\end{tikzpicture}
    \caption{MSLL for data generated from model \eqref{eq:paramm} with varying ranks. Mean and standard deviation plotted from 10 different parts of the data.}
    \label{fig:MSLL}
\end{figure}

Figures \ref{fig:reerr} and \ref{fig:MSLL} show the root mean square error (RMSE) and mean standardized log loss (MSLL) on validation data for different TT-ranks ranging from rank-1 to full-rank ($R=20$).
For data sampled from a model where the ranks are chosen to be low, Algorithm \ref{alg:algi} performs significantly better than the other two methods.
Note, that it is not surprising that Hilbert-GP performs poorly for low ranks since the budget of basis functions to approximate the kernel matrix is very limited.
For the limit of full-rank, the result of Algorithm~\ref{alg:algi} converges to the one of Hilbert-GP. 
This is because in the full-rank case, the size of $\w^{(2)}$ is the same as $\w$, thus $\W_{\setminus2}$ is square.
This case is of course not useful in practice since it does not offer any computational gain compared to Hilbert-GP.

\subsection{Inverse dynamics of a robotic arm}
\label{sec:arm}
\begin{figure}
    \centering
    \input{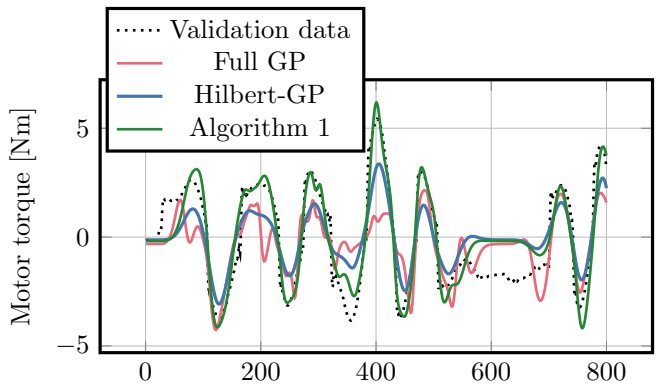}
    \caption{Predicted torques for one joint.}
    \label{fig:invdyn}
\end{figure}

In this experiment, we show that the proposed method works for a real-life data set with a large number of input dimensions, as well as a large number of data points.
We use the benchmark industrial robot data set \citep{weigand2022dataset}, which consists of $N=39\thinspace988$ training points and 3636 test points.
To compute the inverse dynamics of the robotic arm, an 18-dimensional input space (position, velocity, and acceleration for 6 joints) is given together with a 6-dimensional output which are the  motor torques for 6 joints.
We consider just one output in our experiment.
For Hilbert-GP, we use $M=N=39\thinspace988$ and for our method, we use $R=5$ and $M_d=20$ basis functions per dimension, thus $M=20^{18}$.
Table \ref{tab:robot} summarizes the required time for training and prediction on the validation data, as well as the RMSE and MSLL for all used methods.
In addition, Fig.\ \ref{fig:invdyn} shows the RMSE for the first 800 measurements of the validation data set.
Regarding the RMSE, our method performs better than full GP and Hilbert-GP and requires less time.
A possible explanation for this is that the low-rank assumption for the model weights fits the data better than the full GP model.
Regarding the MSLL, our method performs significantly worse than the full GP.
As discussed in section \ref{sec:methods}, through the projection, the posterior covariance matrix of the weights is approximated by a low-rank matrix with a rank equal to $R^2M_d=500$. 
This approximation seems not to be sufficient in this case.

\begin{table}[]
    \caption{Time, RMSE, and MSLL for val. data. }
    \centering
    \begin{tabular}{rlll}
        \toprule 
         & Time [s] & RMSE & MSLL \\ \midrule
         Full GP & 1400 & 1.81 & 7926.01 \\
         Hilbert-GP & 388 & 1.42 & 17805.6 \\
         Our method & 57 & 1.13 & 11621.18 \\
         \bottomrule
    \end{tabular}
    \label{tab:robot}
\end{table}

\section{Conclusion}
In this work, we presented a method for approximate GP regression that uses TNs to project basis functions to a smaller subspace to exploit efficient computations.
We computed a projection matrix in TN format, that projects a parametric model to a smaller subspace, where the distribution of a smaller weight vector is computed.
Finally, we make predictions by projecting the small weight vector back to the original space.
A direction of future work is to explore options to speed up hyperparameter optimization with TNs.
In the reduced-rank GP framework by \cite{HilbertGP}, the approximation leads to inverting a matrix of size $M\times M$ during the hyperparameter optimization. 
Inverting that matrix in TN format efficiently is not straightforward.
Additional future work is to investigate the choice of the ranks of the TN for given data, as well as which of the TT-cores should be chosen for the Bayesian inference.
A limitation of this work is that the computation of the projection matrix is done in a deterministic way.
This causes $\mathbb{V}(\w\mid\y)$ to be underestimated.
An idea is to apply the ALS in a Bayesian framework \citep{menzen2022alternating} to compute a mean and a covariance for all TT-cores and combine the covariance information with, e.g., the unscented transform.

\begin{ack}
This publication is part of the project “Sensor Fusion For Indoor localisation Using The Magnetic Field” with project number 18213 of the research program Veni which is (partly) financed by the Dutch Research Council (NWO).
\end{ack}
\bibliography{ifacconf}

\end{document}